\documentclass[conference]{IEEEtran}
\usepackage{cite}
\usepackage{amsmath,amssymb,amsfonts}
\usepackage{algorithm}
\usepackage[noend]{algorithmic}
\usepackage{graphicx}
\usepackage{textcomp}
\usepackage{xcolor}
\usepackage{dsfont}
\def\BibTeX{{\rm B\kern-.05em{\sc i\kern-.025em b}\kern-.08em
 T\kern-.1667em\lower.7ex\hbox{E}\kern-.125emX}}
\usepackage{amsthm}
\theoremstyle{definition}
\newtheorem{thm}{Theorem}

\newtheorem{lem}{Lemma}
\newtheorem*{cor}{Corollary}
\newtheorem*{conj}{Conjecture}

\begin{document}

\title{RadiX-Net: Structured Sparse Matrices for Deep Neural Networks}

\author{\IEEEauthorblockN{
 Ryan A. Robinett$^1$ and Jeremy Kepner$^{1,2}$
}
\vspace{1ex}
\IEEEauthorblockA{
	$^1$MIT Department of Mathematics, $^2$MIT Lincoln Laboratory Supercomputing Center}
}

\maketitle

\begin{abstract}
The sizes of deep neural networks (DNNs) are rapidly outgrowing the capacity of hardware to store and train them. Research over the past few decades has explored the prospect of sparsifying DNNs before, during, and after training by pruning edges from the underlying topology. The resulting neural network is known as a sparse neural network. More recent work has demonstrated the remarkable result that certain sparse DNNs can train to the same precision as dense DNNs at lower runtime and storage cost. An intriguing class of these sparse DNNs is the X-Nets, which are initialized and trained upon a sparse topology with neither reference to a parent dense DNN nor subsequent pruning. We present an algorithm that deterministically generates RadiX-Nets: sparse DNN topologies that, as a whole, are much more diverse than X-Net topologies, while preserving X-Nets' desired characteristics. We further present a functional-analytic conjecture based on the longstanding observation that sparse neural network topologies can attain the same expressive power as dense counterparts.
\end{abstract}


\section{Introduction}

\let\thefootnote\relax\footnotetext{This material is based in part upon work supported by the NSF under grant number DMS-1312831. Any opinions, findings, and conclusions or recommendations expressed in this material are those of the authors and do not necessarily reflect the views of the National Science Foundation.}

As research in artificial neural networks progresses, the sizes of state-of-the-art deep neural network (DNN) architectures put increasing strain on the hardware needed to implement them \cite{7298594, kepner_exact}. In the interest of reduced storage and runtime costs, much research over the past decade has focused on the sparsification of artificial neural networks \cite{lecun1990optimal,hassibi1993second,srivastava2014dropout,iandola2016squeezenet,DBLP:journals/corr/SrinivasB15,DBLP:journals/corr/HanMD15,7298681,KepnerGilbert2011,kepner2017enabling,kumar2018ibm,kepner2018mathematics}. In the listed resources alone, the methodology of sparsification includes Hessian-based pruning \cite{lecun1990optimal,hassibi1993second}, Hebbian pruning  \cite{srivastava2014dropout}, matrix decomposition \cite{7298681}, and graph techniques \cite{kumar2018ibm,KepnerGilbert2011,kepner2017enabling,kepner2018mathematics}. Yet all of these implementations are alike in that a DNN is initialized and trained, and then edges deemed unnecessary by certain criteria are pruned.

Unlike most strategies for creating sparse DNNs, the X-Net strategy presented in \cite{DBLP:journals/corr/abs-1711-08757} is sparse ``\textit{de novo}"---that is, X-Nets are neural networks initialized upon sparse topologies. X-Nets are observed to train as well on various data sets as their dense counterparts, while exhibiting reduced memory usage \cite{DBLP:journals/corr/abs-1711-08757,alford}. Further, by offering sparse alternatives to fully-connected and convolutional layers---X-Linear and X-Conv layers, respectively---X-Nets exhibit such performance on not only generalized DNN tasks, but also image recognition tasks canonically reserved for convolutional neural networks \cite{7298681}.

X-Net layers are constructed using properties of expander graphs, which give X-Nets the properties of sparsity and path-connectedness (see Mathematical Preliminaries) \cite{DBLP:journals/corr/abs-1711-08757,8186802}. Random X-Linear layers achieve path-connectedness probabilistically, while explicit X-Linear layers, constructed from Cayley graphs, aim to achieve path-connectedness deterministically \cite{DBLP:journals/corr/abs-1711-08757}. 
As an artifact of their construction from Cayley graphs, explicit X-Linear layers are required have the same number of nodes as adjacent layers. This constrains the kinds of X-Net topologies which may be constructed deterministically.

\begin{figure}
 \centering
 \includegraphics[trim={2.5cm 0 2.5cm 0}, clip, width=\columnwidth]{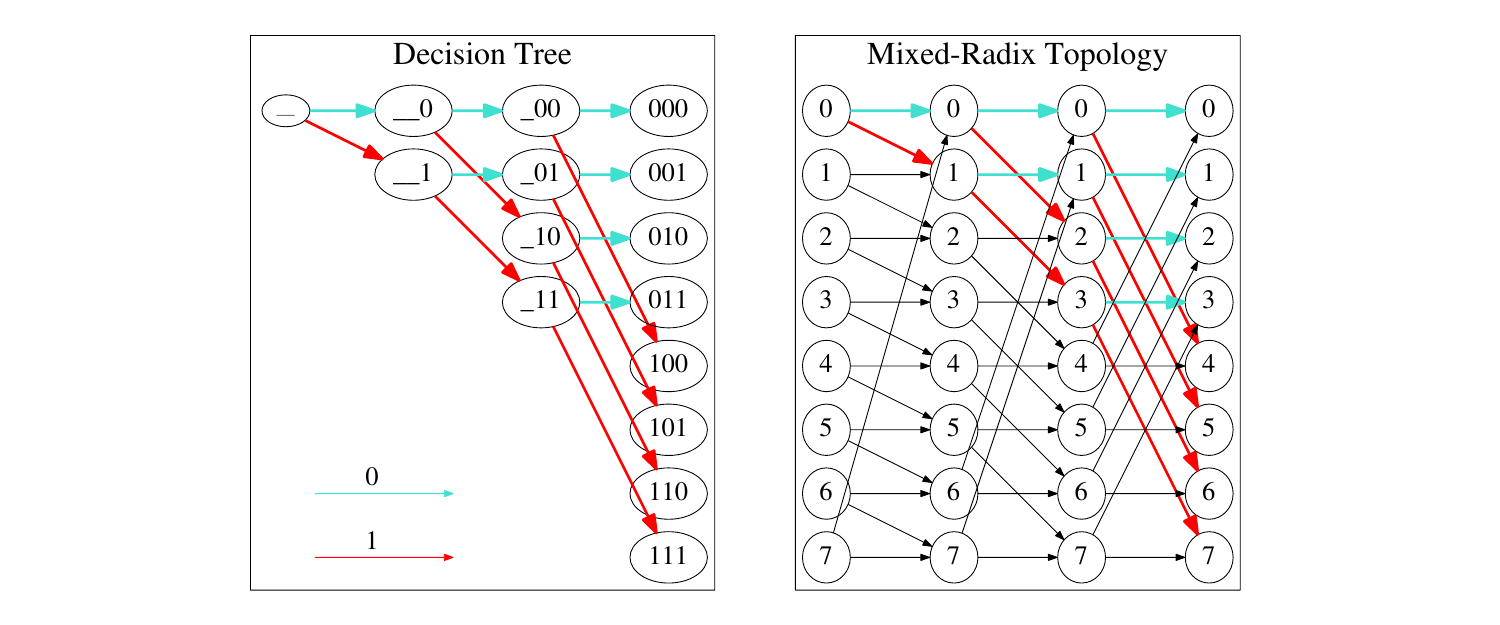}
 \caption{Construction of the mixed-radix topology defined by $\mathcal{N}=(2,2,2)$ using overlapping decision trees. (left) A four-layer binary decision tree. (right) A four-layer mixed radix topology composed of eight offset decision trees.
 }
 \label{fig:Mixed-Radix}
\end{figure}

We propose RadiX-Nets as a new family of \textit{de novo} sparse DNNs that deterministically achieve path-connectedness while allowing for diverse layer architectures. Instead of emulating Cayley graphs, RadiX-Nets achieve sparsity using properties of mixed-radix numeral systems, while allowing for diversity in network topology through the Kronecker product \cite{LOAN200085}. Additionally, RadiX-Nets satisfy symmetry, a property which both guarantees path connectedness and precludes inherent training bias in the underlying sparse DNN architecture.

\begin{figure*}[ht]
 \centering
 \includegraphics[width=7in]{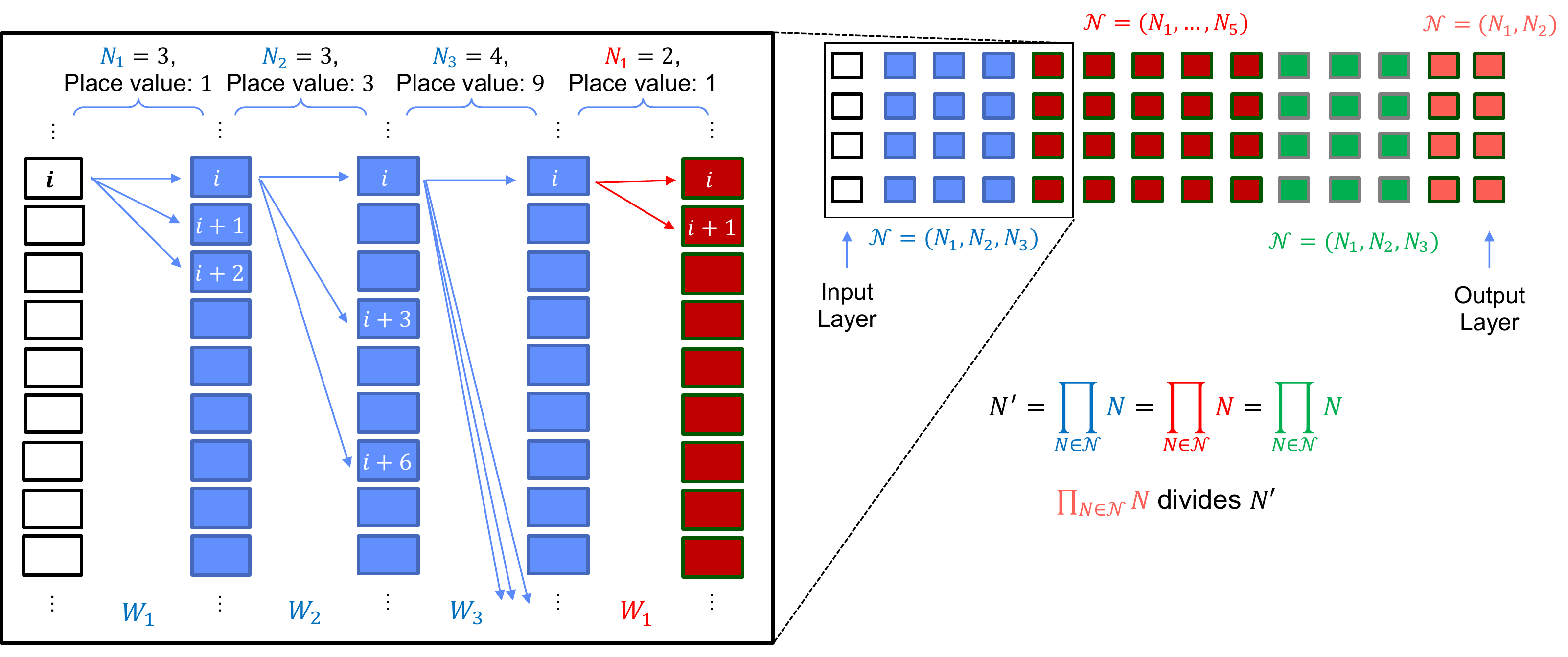}
 \caption{A RadiX-Net \emph{prior} to Kronecker product is a layered graph wherein each layer is a mixed-radix topology. (left) A single mixed-radix topology within a concatenation of mixed-radix topologies, defined by mixed-radix system ${\color{blue}\mathcal{N}=(3,3,4)}$. (top right) A concatenation of the mixed-radix topologies defined by ${\color{blue}\mathcal{N}},{\color{red}\mathcal{N}},{\color{green}\mathcal{N}},$ and ${\color{orange}\mathcal{N}}$. The mixed-radix topologies are concatenated such that the output nodes of one are identified label-wise with the input-nodes of the next. (bottom right) Strict relationships between ${\color{blue}\mathcal{N}},{\color{red}\mathcal{N}},{\color{green}\mathcal{N}},$ and ${\color{orange}\mathcal{N}}$ allow for RadiX-Nets to satisfy sparsity, symmetry, and path-connectedness.}
 \label{fig:Stacking}
\end{figure*}

\section{Mathematical Preliminaries}

\begin{figure*}[ht]
 \centering
 \includegraphics[trim={0 0.5in 0 0.5in}, clip, width=7in]{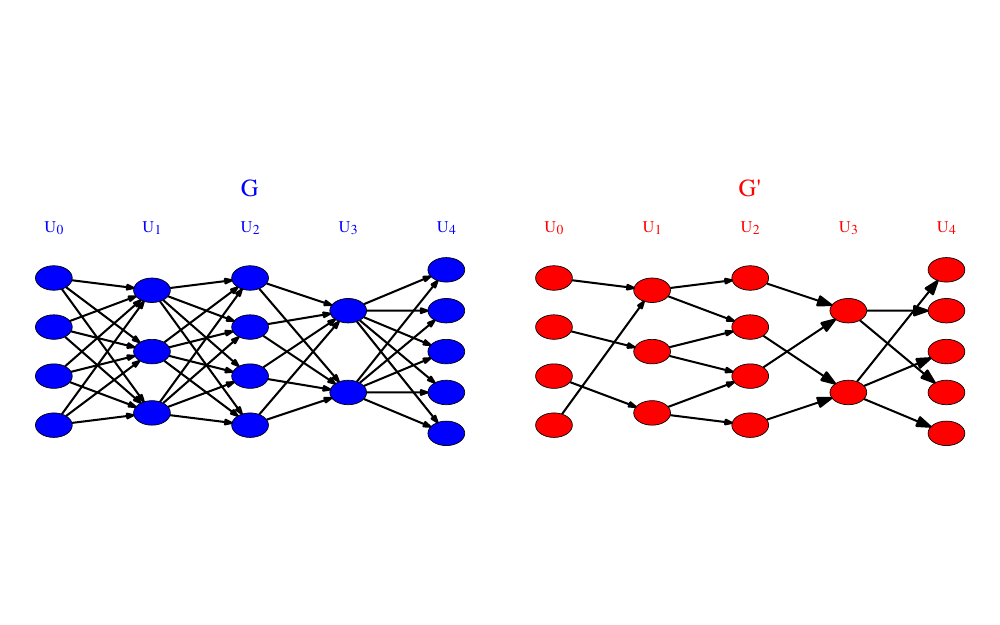}
 \caption{Feedforward neural network topologies (FNNTs) ${\color{blue} G},{\color{red} G^\prime}$ built on the same ordered collection of nodes $\mathcal{U}$. For every ordered collection of nodes, there exists a unique fully-connected FNNT; for $\mathcal{U}$ in this example, this happens to be ${\color{blue} G}$.}
 \label{fig:fnnt}
\end{figure*}

\begin{figure*}[ht]
 \centering
 \includegraphics[trim={0 0.5in 0 0.5in}, clip, width=7in]{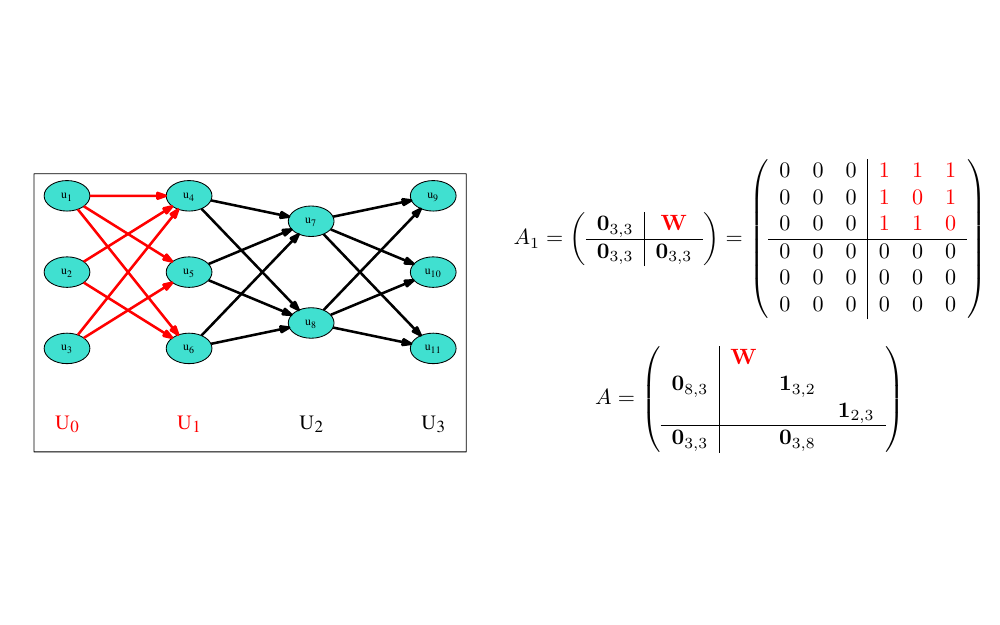}
 \caption{An FNNT $G$, together with the matrices $A_1$ and $A$. $A_1$ is the adjacency matrix of the restriction $G_1$ of $G$ to the the nodes ${\color{red} U_0}\cup{\color{red} U_1}$. By definition, then, ${\color{red} \mathbf{W}}$ is the adjacency submatrix of $G_1$. For both $A_1$ and $A$, the entry at $i,j$ is nonzero if and only if there exists a node from $u_i$ to $u_j$.}
 \label{fig:subadj}
\end{figure*}

\begin{figure*}[ht]
 \centering
 \includegraphics[width=6in]{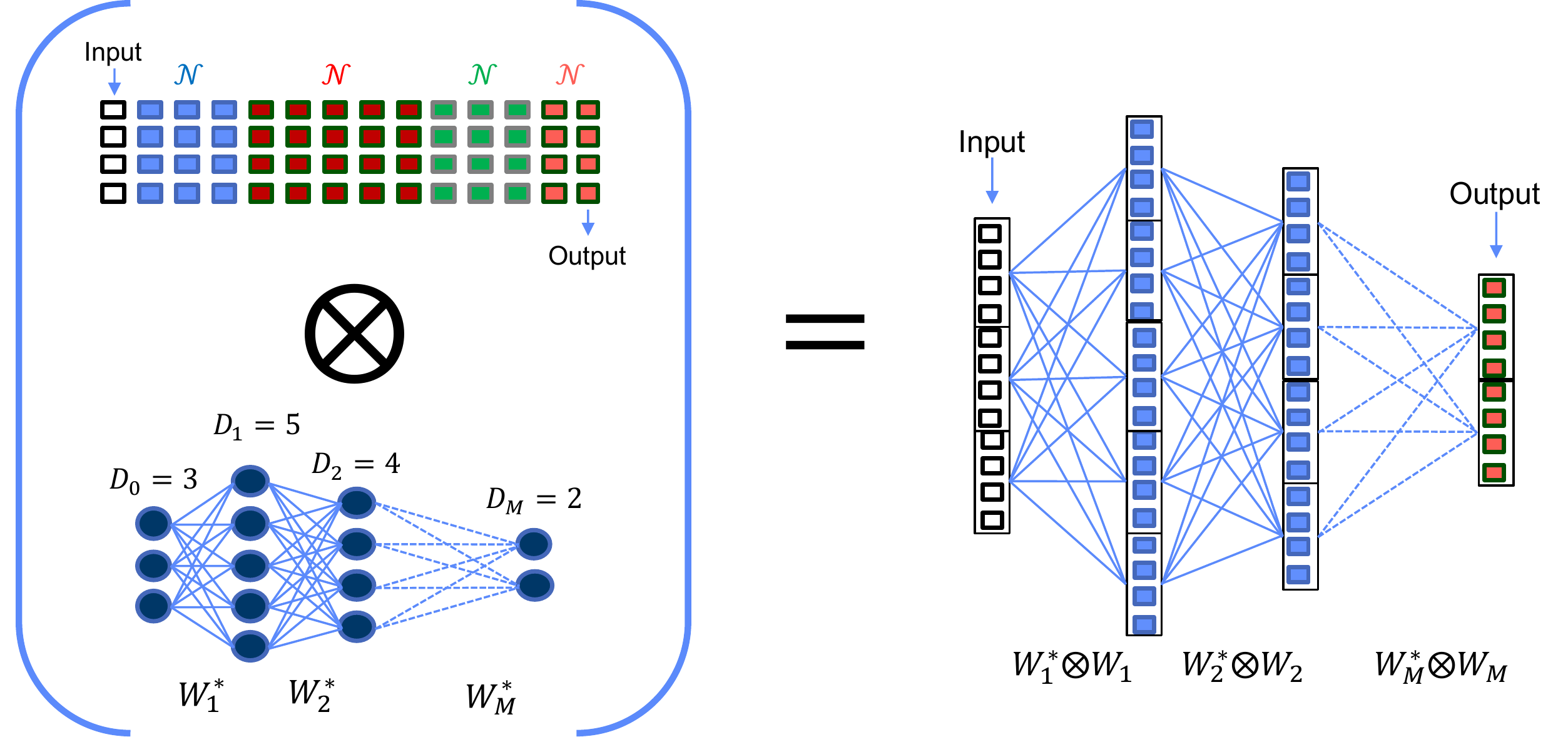}
 \caption{The final step of RadiX-Net construction involves Kronecker products of adjacency submatrices of mixed-radix topologies and adjacency submatrices of an arbitrary dense deep neural network with the same number of layers. The number of vertices in each layer of the dense deep neural networks provides an additional set of parameters by which a wide range of RadiX-Nets can be defined. 
 }
 \label{fig:Kronecker}
\end{figure*}

Understanding RadiX-Nets' graph-theoretic construction and underlying mathematical properties requires defining a few concepts. RadiX-Nets are composed of sub-nets that are herein referred to as mixed-radix topologies. Mixed-radix topologies are based on properties of mixed-radix number systems, and can be constructed from overlapping decision trees (see Figure~\ref{fig:Mixed-Radix}). A mixed-radix numeral system is the sole parameter used to uniquely specify a mixed-radix topology. Mixed-radix topologies are a kind of feedforward neural net topology (FNNT), which is a layered graph wherein all vertices in one layer point only to some number of vertices in the next. The adjacency matrix of an FNNT is uniquely defined by the adjacency submatrices corresponding to each of its layers. Essentially, RadiX-Net topologies are constructed from Kronecker products of mixed-radix adjacency submatrices and dense DNN adjacency submatrices (see Figure~\ref{fig:Kronecker}). The main properties of interest in RadiX-Nets are path-connectedness---which ensures each output depends upon all inputs---and symmetry, which ensures that there is the same number of paths between each input and output.

\emph{Mixed-Radix Numeral System}:
Let $\mathcal{N}=(N_1,\ldots,N_L)$ be an ordered set of $L$ integers greater than 1. Let $N^\prime=\prod_{i=1}^LN_i$. All such $\mathcal{N}$ implicitly define a numeral system which bijectively represents all integers in $\{0,\ldots,N^\prime-1\}$. That is, the set of ordered sets
$$\big\{(n_1,\ldots,n_L)\mid n_i\in\{0,\ldots,N_i-1\}\big\}$$
maps bijectively to $\{0,\ldots,N^\prime-1\}$ by the map
$$(n_1,\ldots,n_L)\longleftrightarrow\sum_{i=1}^L\left(n_i\prod_{j=1}^{i-1}N_j\right).$$
Mixed-radix numeral systems arise naturally in numerous graph-theoretic constructions, such as decision trees (see Figure~\ref{fig:Mixed-Radix}).

\emph{Feedforward Neural Net Topology (FNNT)}: 
An FNNT $G$ with $n+1$ layers of nodes---including input and output layers---is an $(n+1)$-partite directed graph with independent components $U_0,\ldots,U_n$ satisfying the constraints that
\begin{itemize}
\item if there exists an edge from $u\in U_i$ to $v\in U_j$, then $j=i+1$, and
\item the out-degree of $u\in U_i$ is nonzero for all $i<n$.
\end{itemize}

\emph{Adjacency Submatrix of an FNNT}:
Say $G$ is an FNNT. Let $G_i$ be the restriction of $G$ to the set of nodes $U_{i-1}\cup U_i$ and the set of edges from $U_{i-1}$ to $U_i$ in $G$. We define $m_i=\lvert U_{i-1}\rvert$ and $n_i=\lvert U_i\rvert$ for all $i$. Up to a permutation of indices, the adjacency matrix of $G_i$ is of the form
$$\left(\begin{array}{c | c}
\mathbf{0}_{m_i,m_i} & \mathbf{W}_i \\
\hline
\mathbf{0}_{n_i,m_i} & \mathbf{0}_{n_i,n_i} \\
\end{array}\right)$$
for some $\mathbf{W}_i$, where $\mathbf{0}_{a,b}$ is the $a\times b$ matrix of zeros. We refer to $\mathbf{W}_i$ as the adjacency submatrix of the restriction $G_i$.

Conversely, say that an ordered set $\mathcal{W}=(\mathbf{W}_1,\ldots,\mathbf{W}_n)$ of matrices is such that
\begin{itemize}
\item the only nonzero entries of $\mathbf{W}_i$ are ones for all $i$, and
\item no column of $\mathbf{W}_i$ is the zero vector.
\end{itemize}
If the number of columns in $\mathbf{W}_{i-1}$ equals the number of rows in $\mathbf{W}_i$ for all $i\in\{1,\ldots,n\}$, then $\mathcal{W}$ defines a unique FNNT with $n+1$ layers of nodes.

\emph{Path-Connectedness}:
We define path-connectedness as follows: let $G$ be an FNNT with $n+1$ layers of nodes. $G$ is path-connected if, for every $u\in U_0$ and every $v\in U_n$, there exists a path from $u$ to $v$.

\emph{Symmetry}:
We define symmetry as follows: let $G$ be an FNNT with $n+1$ layers of nodes. $G$ is symmetric if there exists a positive integer $m$ such that, for all $u\in U_0$ and all $v\in U_n$, there exist exactly $m$ paths from $u$ to $v$. If $G$ is symmetric, it is path-connected. If $G$ has adjacency matrix $A$, then $G$ satisfies symmetry if and only if, up to some permutation of $A$,
$$A^n=\left(\begin{array}{c|c}
    \mathbf{0}_{n,M-n} & m\mathbf{1}_{n,n} \\
    \hline
    \mathbf{0}_{M-n,M-n} & \mathbf{0}_{M-n, n}
\end{array}\right),$$
where $M$ is the number of nodes in $G$, $\mathbf{1}_{a,b}$ is the $a\times b$ matrix of ones, and $m$ is some positive integer.

\emph{Density of an FNNT}
An ordered collection $(U_0,\ldots,U_n)$ of sets of nodes implicitly defines a unique, fully-connected DNN topology---namely, the FNNT such that, for all $i\in\{1,\ldots,n\}$, there exists an edge from $u$ to $v$ for all $u\in U_{i-1}$ and all $v\in U_i$. The number of edges in this DNN topology is equal to $\sum_{i=1}^n\lvert U_{i-1}\rvert\lvert U_i\rvert$. We define the density of an FNNT $G$ as the ratio of the number of edges in $G$ to the number of edges in the DNN topology defined by the ordered set of independent components of $G$. By this construction, the highest possible density of an FNNT is one, while the lowest is $\frac{\sum_{i=1}^n\lvert U_{i-1}\rvert}{\sum_{i=1}^n\lvert U_{i-1}\rvert\lvert U_i\rvert}$.

\section{RadiX-Net Topologies}

\begin{figure}
 \centering
 \includegraphics[trim={0 1.8cm 0 1.5cm}, clip, width=8cm]{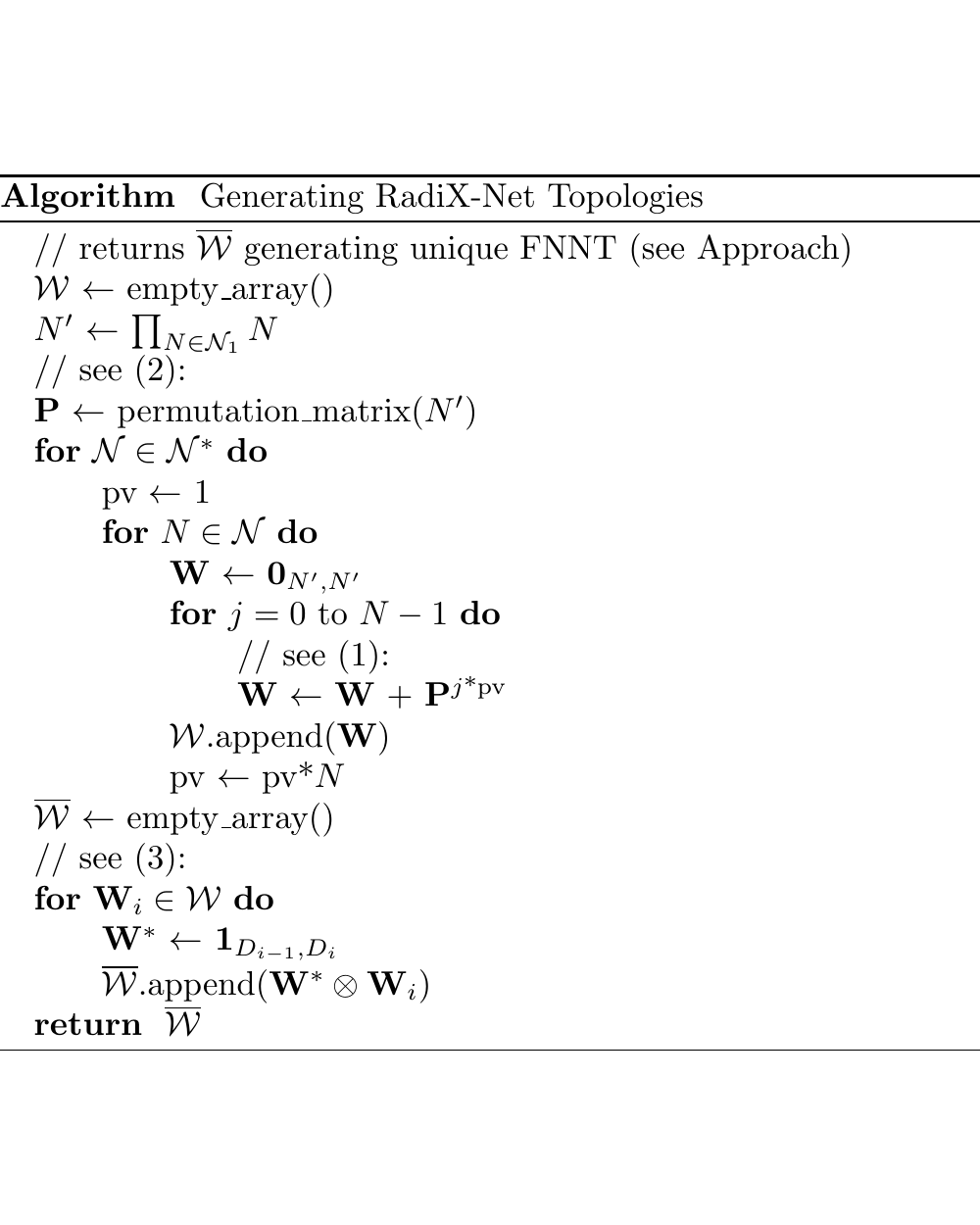}
 \caption{An algorithm for generating the RadiX-Net topology defined by list $\mathcal{N}^*=(\mathcal{N}_1,\ldots,\mathcal{N}_M)$ of mixed-radix numeral systems and list $\mathcal{D}=(D_0,\ldots,D_{\overline{M}})$ of positive integers.}
 \label{fig:pseudo}
\end{figure}

\subsection{Constructing RadiX-Net Topologies}

We construct RadiX-Net topologies using mixed-radix topologies as building blocks, as motivated by Figure \ref{fig:Stacking}.

\emph{Mixed-Radix Topologies:}
Let $L$ be a positive integer, and let $\mathcal{N}=(N_1,\ldots,N_L)$, where $N_i$ is an integer greater than 1 for all $i$. Let $N^\prime=\prod_{N\in\mathcal{N}}N$, and let $U_i$ be a set of $N^\prime$ nodes---with labels $0,\ldots,N^\prime-1$---for all $i\in\{0,\ldots,L\}$. For all $i\in\{1,\ldots,L\}$, we create edges from node $j$ in $U_{i-1}$ to node $j+n\prod_{j=1}^{i-1}N_j\textmd{ (mod }N^\prime)$ in $U_i$ for all $n\in\{0,\ldots,N_i-1\}$. Let $\mathbf{W}_i$ be the adjacency submatrix defining the edges from $U_{i-1}$ to $U_i$. By construction, we have that
\begin{equation} \label{circshift}
\mathbf{W}_i = \sum_{j=0}^{N_i-1}\mathbf{P}^{j\nu_i},
\end{equation}
where $\nu_i=\prod_{k=1}^{i-1}N_k$ and $\mathbf{P}$ is the permutation matrix
\begin{equation} \label{perm}
\left(\begin{array}{ccc|c}
0 & \ldots & 0 & 1 \\
\hline
\ & \ & \ & 0 \\
\ & \mathbf{I}_{N^\prime-1} & \ & \vdots \\
\ & \ & \ & 0
\end{array}\right),
\end{equation}
$\mathbf{I}_n$ being the $n\times n$ identity matrix. We refer to the resulting graph as the mixed-radix topology induced by $\mathcal{N}$.

\emph{RadiX-Net Topologies:}
Here, we formally construct RadiX-Net topologies using mixed-radix topologies, adjacency submatrices, and the Kronecker product, as motivated by Figure \ref{fig:Kronecker}. For an informal programmatic construction, see Figure \ref{fig:pseudo}.

RadiX-Net topologies are uniquely defined by an ordered set $\mathcal{N}^*=(\mathcal{N}_1,\ldots,\mathcal{N}_M)$ of mixed-radix numeral systems $\mathcal{N}_i=(N_1^i,\ldots,N_{L_i}^i)$ together with an ordered set $\mathcal{D}$ of positive integers. We require that
\begin{enumerate}
 \item there exists a positive integer $N^\prime$ such that $N^\prime=\prod_{N\in\mathcal{N}_i}N$ for all $i\in\{1,\ldots,M-1\}$, and
 \item $\prod_{N\in\mathcal{N}_M}N$ divides $N^\prime$.
\end{enumerate}
Let $\overline{M}=\sum_{i=1}^{M}L_i$, the total number of radices in $\mathcal{N}^*$; we further require that $\mathcal{D}=(D_0,\ldots,D_{\overline{M}})$ consist of $\overline{M}+1$ integers satisfying $D_i\ll N^\prime$ for all $i$.

We construct a RadiX-Net $G$ using $\mathcal{N}^*$ and $\mathcal{D}$ as follows: let $G_i$ be the mixed-radix topology induced by $\mathcal{N}_i$. Identifying the output nodes of $G_i$ with the input nodes of $G_{i+1}$ creates an $\overline{M}$-layer FNNT with ordered set $\mathcal{W}=(\mathbf{W}_1,\ldots,\mathbf{W}_{\overline{M}})$ of adjacency submatrices of the form (\ref{circshift})\textsuperscript{$\dagger$}\footnote{\textsuperscript{$\dagger$}We refer to such an FNNT as an \textit{extended mixed-radix topology} (see Appendix).}. Similarly, $\mathcal{D}$ implicitly defines a unique dense DNN topology $H$ on an ordered collection $U_0,\ldots,U_{\overline{M}}$ of nodes satisfying $\lvert U_i\rvert=D_i$. The ordered set of adjacency matrices of $H$ is $\mathcal{W}^*=(\mathbf{W}_1^*,\ldots,\mathbf{W}_{\overline{M}}^*)$, where $\mathbf{W}_i^*$ is the $D_{i-1}\times D_i$ matrix of ones. We define $G$ as the unique FNNT defined by
\begin{equation} \label{krons}
\overline{\mathcal{W}}=(\mathbf{W}_1^*\otimes\mathbf{W}_1,\ldots,\mathbf{W}_{\overline{M}}^*\otimes\mathbf{W}_{\overline{M}})
\end{equation}
(see Mathematical Preliminaries).

Mixed-radix and RadiX-Net topologies satisfy symmetry, and therefore path-connectedness. Proofs for this assertion, as well as the number of paths from any node $u$ in the input layer to a node $v$ in the output layer for each family of topologies, can be found in the Appendix.

\subsection{Asymptotic Sparsity of RadiX-Nets}

\begin{figure}
 \centering
 \includegraphics[width=\columnwidth]{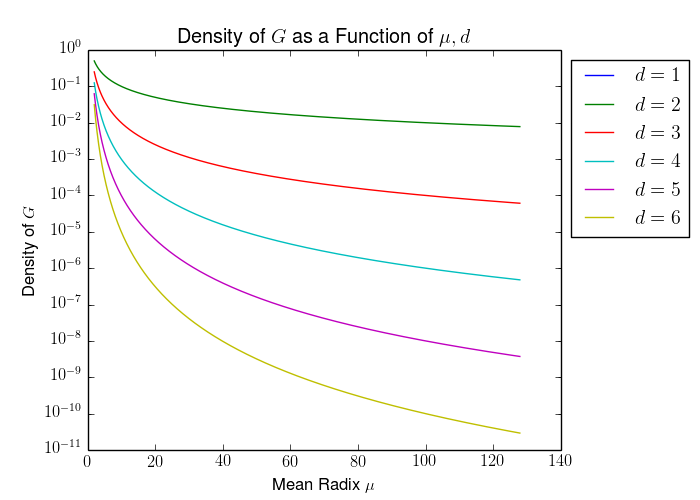}
 \caption{If $G$ is the RadiX-Net topology defined using $\{\overline{N}_i\},\{D_i\}$ as defined in (\ref{eqn:density}), and $\{\overline{N}_i\}$ has sufficiently small variance, then the density of a $G$ varies as a function of $\mu$ and $d$ (see (\ref{eqn:approx})).}
 \label{fig:sparsity}
\end{figure}

Say $G$ is the RadiX-Net topology generated by $\mathcal{N}^*=(\mathcal{N}_1,\ldots,\mathcal{N}_M),\mathcal{D}=(D_0,\ldots,D_{\overline{M}})$. Further say $\mathcal{N}_i=(N_{i,1},\ldots,N_{i,L_i})$ for all $i$, and let $N^\prime$ be the integer satisfying $N^\prime=\prod_{N\in\mathcal{N}_i}N$ for all $i\in\{1,\ldots,M-1\}$. If we define
$$(\overline{N}_1,\ldots,\overline{N}_{\overline{M}}):=(N_{1,1},\ldots,N_{1,L_1},N_{2,1},\ldots,N_{M,L_M}),$$
then the density $\Delta_G$ of $G$ is given by
\begin{equation}\label{eqn:density}
    \Delta_G=\left(\frac{1}{N^\prime}\right)\left(\frac{\sum_{i=1}^{\overline{M}}\overline{N}_iD_{i-1}D_i}{\sum_{i=1}^{\overline{M}}D_{i-1}D_i}\right).
\end{equation}
Let $\mu$ be the mean value of $\{\overline{N}_i\}$. When $\{\overline{N}_i\}$ has sufficiently small variance, it follows immediately from (\ref{eqn:density}) that
\begin{equation}\label{eqn:mu}
    \Delta_G\approx\frac{\mu}{N^\prime}.
\end{equation}
This implies that when $\{\overline{N}_i\}$ has small variance, the sparsity of $G$ is negligibly affected by $\{D_i\}$.

We define $d=\log_{\mu}N^\prime$. For sufficiently small variance of the $\overline{N}_i$, we can assume that $d$ is approximately equal to some integer, with which we can write
\begin{equation}\label{eqn:approx}
    \Delta_G\approx\frac{1}{\mu^{d-1}}.
\end{equation}
Concretely, $\mu$ corresponds to the average radix of each mixed-radix numeral system used to construct $G$, and $d$ corresponds to the number of radices used to construct each mixed-radix numeral system\textsuperscript{$\ddagger$}\footnote{\textsuperscript{$\ddagger$}Per bullet 2) in Section III.A, this excludes the last mixed-radix numeral system.}\textsuperscript{*}\footnote{\textsuperscript{*}Note that this assumption is contingent on $\{N_i\}$ having sufficiently small variance.}. The effect of $\mu$ and $d$ on the sparsity of $G$ is shown in Figure \ref{fig:sparsity}.

\section{Conclusions \& Future Work}

This paper presents the RadiX-Net algorithm, which deterministically generates sparse DNN topologies that, as a whole, are much more diverse than X-Net topologies while preserving X-Net's desired characteristics. In a related effort, benchmarking RadiX-Net performance in comparison to X-Net, dense DNN, and other neural network implementations can be found in \cite{alford}. Furthermore, RadiX-Net is used in \cite{wang} to construct a neural net simulating the size and sparsity of the human brain.

Prabhu \textit{et al.} and Alford \textit{et al.} come at the end of a long history of sparse neural network research\cite{lecun1990optimal,hassibi1993second,srivastava2014dropout,iandola2016squeezenet,DBLP:journals/corr/SrinivasB15,DBLP:journals/corr/HanMD15,7298681,KepnerGilbert2011,kepner2017enabling,kumar2018ibm,kepner2018mathematics,DBLP:journals/corr/abs-1711-08757,alford}. This collective body mutually corroborates the following assertion: Sparse neural networks can train to the same arbitrary degree of precision as their dense counterparts. While the reduced training time of sparse neural nets can be attributed to having fewer parameters, there is no intuitive reason as to why sparse networks should demonstrate the same expressive power---as some have put it---as dense counterparts.

Na\"ively, should sparse networks have the same expressive power as dense networks, dense and pruned networks would be obsolete, as \textit{de novo} sparse networks achieve the expressive power of both while exceeding the training speed of both. Because the corpus of research in sparse networks seems unanimous on the subject, it would behoove the field to become more objective about what is meant when discussing expressive power, as is done in \cite{NIPS2017_7203,khrulkov2018expressive,pmlr-v70-raghu17a}. As demonstrated by \cite{Cybenko1989}, functional analysis provides a powerful language with which to describe the abilities and limitations of neural networks rigorously. In Section IV.B, we present a functional-analytic conjecture based on the mentioned experimental findings, which the authors intend to prove at a later date. Posing and proving such conjectures would direct future research in artificial neural networks more prudently than would experimental results alone.

\subsection{Preliminaries for Conjecture}

\begin{figure*}
 \centering
 \includegraphics[width=7in]{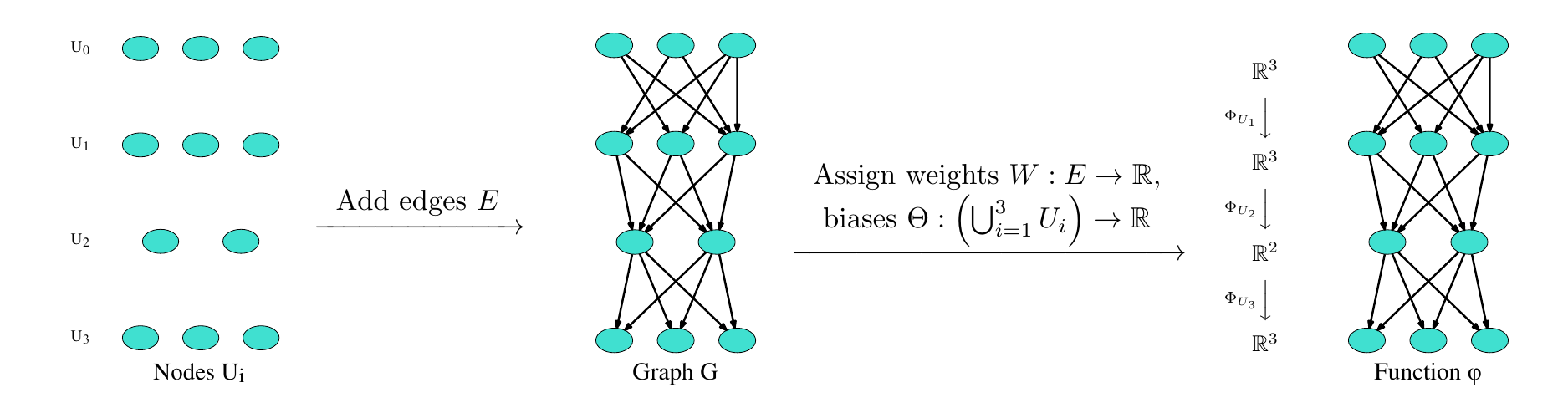}
 \caption{Treating all activation functions as equal, a feedforward neural network (FNN) $\mathcal{G}$ is uniquely determine by an finite ordered collection of nodes $\mathcal{U}$, a set of edges $E$ that makes $\mathcal{U}$ into an FNNT, a weights function $W:E\to\mathbb{R}$, and a bias function $\Theta:\left(\bigcup_{i\neq0}U_i\right)\to\mathbb{R}$. Given $m$ is the highest index of the $U_i\in\mathcal{U}$, $\mathcal{G}$ induces the unique function $\varphi:\mathbb{R}^{\lvert U_0\rvert}\to\mathbb{R}^{\lvert U_m\rvert}$ defined by $\varphi=\Phi_m\circ\ldots\circ\Phi_0$, where the element of $\Phi_i(\vec{x})$ corresponding to $u\in U_i$ is equal to $\varphi_u(\vec{x})$ (see (\ref{subPhi})).}
 \label{fig:layering}
\end{figure*}

The most sturdy theoretical ground upon which artificial neural nets stand is Cybenko's Universality Theorem. Though the original statement of the theorem is stronger than the corollary below, this corollary captures the significance of the Universality Theorem in the field of artificial neural networks.
\begin{cor}
Let $\sigma:\mathbb{R}\to\mathbb{R}$ be a continuous function such that $\lim_{t\to\infty}\sigma(t)=1$ and $\lim_{t\to-\infty}\sigma(t)=0$ (let us call this function sigmoidal). Further, let $\mathcal{C}_n$ be the space of continuous functions on $I_n=[0,1]^n$ with metric topology defined by supremum norm $d(f,g)=\sup_{\vec{x}\in I_n}\lvert f(\vec{x})-g(\vec{x})\rvert$. Lastly, let $S$ be the set of functions of the form
\begin{equation} \label{sigmoids}
G(\vec{x})=\sum_{j=1}^N\alpha_j\sigma\left(\vec{y}_j^T\vec{x}+\theta_j\right),
\end{equation}
where $N$ is a natural number, $\alpha_j$ and $\theta_j$ are real numbers, and $\vec{y}$ is an element of $\mathbb{R}^n$. The set $S$ is dense $\mathcal{C}_n$. $\qedsymbol$
\end{cor}
We adopt some of the language of this corollary to make our conjecture connect more immediately to the literature.

Let $\sigma$, $I_n$, $\mathcal{C}_n$, and $d$ be as defined above. We define a feedforward neural network (FNN) as an FNNT $G$, with set of edges $E$, together with a map $W:E\to\mathbb{R}$ assigning a weight $w$ to each edge and a map $\Theta:\bigcup_{i=1}^mU_i\to\mathbb{R}$---where $m$ is the number of non-input layers in $G$---assigning a bias $\theta$ to each non-input node. We associate with each FNN $\mathcal{G}$ the unique map $\varphi:\mathbb{R}^{\lvert U_0\rvert}\to\mathbb{R}^{\lvert U_m\rvert}$ defined by the following:
\begin{itemize}
\item let $\tilde{E}:\bigcup_{i=1}^mU_i\to E$ map each node $u$ to the set of edges going into $u$;
\item for all $u_i\in U_0$, let $\varphi_{u_i}(x_1,\ldots,x_{\lvert U_0\rvert})=x_i$;
\item for all $i\in\{1,\ldots,m\}$ and for all $v\in U_i$, let
\begin{equation} \label{subPhi}
\varphi_v(\vec{x})=\sigma\left(\Theta(v)+\sum_{(u,v)\in\tilde{E}(v)}W(u,v)\varphi_u(\vec{x})\right);
\end{equation}
\item assuming $U_m=\{u_1,\ldots,u_{\lvert U_m\rvert}\}$, we define
\begin{equation} \label{eqn:phi}
\varphi(\vec{x})=\left(\varphi_{u_1}(\vec{x}),\ldots,\varphi_{u_{\lvert U_m\rvert}}(\vec{x})\right).
\end{equation}
\end{itemize}

Let $\mathcal{U}=(U_0,U_1,\ldots)$ be an infinite ordered collection of finite sets of nodes such that $\lvert U_0\rvert=n$. Let $\mathfrak{D}$ be the unique fully-connected FNNT on $\mathcal{U}$, and let $\mathcal{S}$ be some sparse FNNT on $\mathcal{U}$ satisfying symmetry. We define $\mathfrak{D}_N$ and $\mathcal{S}_N$ as the unique FNNTs constructed by restricting $\mathfrak{D}$ and $\mathcal{S}$, respectively, to the set of nodes $\bigcup_{i=0}^NU_i$, introducing a new node $v$, and creating and edge from $u$ to $v$ for all $u\in U_N$. Finally, let $\mathbb{D}_N$ and $\mathbb{S}_N$ be the sets of continuous functions which can be represented as FNNs on $\mathfrak{D}_N$ and $\mathcal{S}_N$, respectively.

\subsection{Functional-Analytic Conjecture}

Due to the findings of Prabhu \textit{et al.}, Alford \textit{et al.}, and others, we are convinced that \textit{de novo} sparse neural network topologies exhibit the same expressive power of fully-connected DNN topologies in the following way.

\begin{conj}
For all $\mathbb{X}\subset\mathcal{C}_n$, we define
\begin{equation} \label{delta}
\delta(\mathbb{X})=\sup_{f\in\mathcal{C}_n}\left[\inf_{g\in\mathbb{X}}\left(d(f,g)\right)\right].
\end{equation}
If $\delta(\mathbb{D}_N)$ is in $O(N^{-p})$ for some $p$, then $\delta(\mathbb{S}_N)$ is also in $O(N^{-p})$.\hfill$\qedsymbol$
\end{conj}

\section*{Acknowledgment}
The authors wish to acknowledge the following individuals for their contributions and support: Simon Alford, Alan Edelman, Vijay Gadepally, Chris Hill, Hayden Jananthan, Lauren Milechin, Richard Wang, and the MIT SuperCloud team.

\bibliography{RadiX-Net}
\bibliographystyle{ieeetr}

\section*{Appendix}

For purposes of simplifying Theorem \ref{thm:radix}, we use the following two lemmas. Lemma \ref{lem:emr} discusses \textit{extended mixed-radix topologies}, which we define as RadiX-Net topologies generated by $\mathcal{N}^*,\mathcal{D}=(D_0,\ldots,D_{\overline{M}})$ satisfying $D_i=1$ for all $i$.

\begin{lem}\label{lem:mr}
    Mixed-radix topologies satisfy symmetry, and the number of paths from an input node $u$ to an output node $v$ is one.
    \begin{proof}
        This follows directly from the definition of a mixed-radix numeral system.
    \end{proof}
\end{lem}

\begin{lem}\label{lem:emr}
    Let $G$ be the extended mixed-radix (EMR) topology defined by some $\mathcal{N}^*=(\mathcal{N}_1,\ldots,\mathcal{N}_M)$ satisfying the RadiX-Net constraints (see Section III: RadiX-Net Topologies). $G$ satisfies symmetry, and the number of paths from an input node $u$ to an output node $v$ is $(N^\prime)^{M-1}$, where $N^\prime$ is the integer satisfying $N^\prime=\prod_{N\in\mathcal{N}_i}N$ for all $i\in\{1,\ldots,M-1\}$.
    \begin{proof}
        We show this by induction. Say that, for some positive integer $M$, all EMR topologies $G$ defined by some $\mathcal{N}^*=(\mathcal{N}_1,\ldots,\mathcal{N}_M)$ satisfy symmetry. Let $\mathcal{N}_+^*=(\mathcal{N}_1,\ldots,\mathcal{N}_M,\mathcal{N}_{M+1})$ for some $\mathcal{N}_{M+1}$ satisfying the RadiX-Net constraints, and let $G_+$ be the EMR topology induced by $\mathcal{N}_+^*$. Recall that $G_+$ is formed from the disjoint union of the MR topologies $G_i$ (generated by $\mathcal{N}_i$) by identifying $U_{i-1,L_{i-1}}$ and $U_{i,0}$ for all $i$ (here, $U_{i,L_i}$ and $U_{i,0}$ simply refer to the output and input layers, respectively, of $G_i$). Because $G_{M+1}$ is an MR topology, Lemma \ref{lem:mr} guarantees that there exists exactly one path from $u$ to $v$ for all $u\in U_0^{M+1}$ and all $v\in U_{L_{M+1}}^{M+1}$. By hypothesis, for some positive integer $m$, there exist exactly $m$ paths from $\tilde{u}\in U_{1,0}$ to $\tilde{v}\in U_{M,L_M}$ for all such $\tilde{u},\tilde{v}$. Because $U_{M,L_M}$ and $U_{M+1,0}$ are identified, this implies that for every path from $\tilde{u}\in U_{1,0}$ to $\tilde{v}\in U_{M,L_M}$, there exists exactly one path from $\tilde{u}$ to $v\in U_{M+1,L_{M+1}}$ which passes through $\tilde{v}$. Further, because there are $\lvert U_{M+1,0}\rvert$ such $\tilde{v}$, there exist exactly $m\lvert U_{M+1,0}\rvert$ paths from $\tilde{u}$ to $v$ for all choices of $\tilde{u},v$. By induction from the case $M=1$ (i.e. Lemma \ref{lem:mr}), $G_+$ satisfies symmetry, and $m=\prod_{i=2}^{M}\lvert U_{i,0}\rvert=(N^\prime)^{M-1}$.
    \end{proof}
\end{lem}

\begin{thm}\label{thm:radix}
    Let $G$ be the RadiX-Net topology defined by some $\mathcal{N}^*=(\mathcal{N}_1,\ldots,\mathcal{N}_M),\mathcal{D}=(D_0,\ldots,D_{\overline{M}})$ satisfying the RadiX-Net constraints. We order the layers $\overline{U}_0,\ldots,\overline{U}_{\overline{M}}$ of $G$ in the natural way, where $\overline{U}_0$ and $\overline{U}_{\overline{M}}$ are the input and output layers, respectively, of $G$. $G$ satisfies symmetry, and the number of paths from input node $u$ to output node $v$ is given by $(N^\prime)^{\overline{M}-1}\left(\prod_{i=1}^{\overline{M}-1}D_i\right)$, where $N^\prime$ is the integer satisfying $N^\prime=\prod_{N\in\mathcal{N}_i}N$ for all $i\in\{1,\ldots,M-1\}$.
    \begin{proof}
        Let $\mathbf{A}$ be the adjacency matrix of $G$, and let $\mathbf{W}^*_i,\mathbf{W}_i$ be as defined in (\ref{krons}). We define $\kappa=N^\prime\sum_{i=0}^{\overline{M}}D_i$, $\alpha=N^\prime D_0$, and $\beta=N^\prime D_{\overline{M}}$.
        Up to a permutation, $\mathbf{A}$ is of the form
        \begin{equation} \label{bigA}
        \left(\begin{array}{c | c c c}
        \ & \mathbf{W}^*_1\otimes\mathbf{W}_1 & \ & \ \\
        \mathbf{0}_{\kappa-\beta,\alpha} & \ & \ddots & \ \\
        \ & \ & \ & \mathbf{W}^*_{\overline{M}}\otimes\mathbf{W}_{\overline{M}} \\
        \hline
        \mathbf{0}_{\beta,\alpha} & \ & \mathbf{0}_{\beta,\kappa-\alpha} & \
        \end{array}\right).
        \end{equation}
        Therefore, the following statements hold.
        \begin{align*}
        \mathbf{A}^{\overline{M}}&=\left(\begin{array}{c | c}
        \mathbf{0}_{\alpha,\kappa-\beta} & \prod_{i=1}^{\overline{M}}(\mathbf{W}^*_i\otimes\mathbf{W}_i) \\
        \hline
        \mathbf{0}_{\kappa-\alpha,\kappa-\beta} & \mathbf{0}_{\kappa-\alpha,\beta}
        \end{array}\right) \\
        &=\left(\begin{array}{c | c}
        \mathbf{0}_{\alpha,\kappa-\beta} & \left(\prod_{i=1}^{\overline{M}}\mathbf{W}^*_i\right)\otimes\left(\prod_{i=1}^{\overline{M}}\mathbf{W}_i\right) \\
        \hline
        \mathbf{0}_{\kappa-\alpha,\kappa-\beta} & \mathbf{0}_{\kappa-\alpha,\beta}
        \end{array}\right)
        \end{align*}
        The deduction above is consequent of the mixed-product property of the Kronecker product\cite{LOAN200085}. It is easy to show that
        \begin{equation} \label{prodB}
        \prod_{i=1}^{\overline{M}}\mathbf{W}^*_i=\left(\prod_{i=1}^{\overline{M}-1}D_i\right)\left(\mathbf{1}_{D_0,D_{\overline{M}}}\right),
        \end{equation}
        where $\mathbf{1}_{a,b}$ is the $a\times b$ matrix of ones. By Lemma \ref{lem:emr}, it holds that
        \begin{equation} \label{prodW}
        \prod_{i=1}^{\overline{M}}\mathbf{W}_i=\left(N^\prime\right)^{\overline{M}-1}\left(\mathbf{1}_{N^\prime,N^\prime}\right).
        \end{equation}
        Therefore,
        $$\mathbf{A}^{\overline{M}}=\left(\begin{array}{c | c}
        \mathbf{0}_{\alpha,\kappa-\beta} & \left(N^\prime\right)^{\overline{M}-1}\left(\prod_{i=1}^{\overline{M}-1}D_i\right)\left(\mathbf{1}_{\alpha,\beta}\right) \\
        \hline
        \mathbf{0}_{\kappa-\alpha,\kappa-\beta} & \mathbf{0}_{\kappa-\alpha,\beta}
        \end{array}\right).$$
        So $G$ satisfies symmetry, and for all input nodes $u$ and output nodes $v$, there exist exactly $\left(N^\prime\right)^{\overline{M}-1}\left(\prod_{i=1}^{\overline{M}-1}D_i\right)$ paths from $u$ to $v$.
    \end{proof}
\end{thm}

\end{document}